\documentclass[final]{cvpr}

\usepackage{times}
\usepackage{epsfig}
\usepackage{graphicx}
\usepackage{amsmath}
\usepackage{amssymb}
\usepackage{multirow}
\usepackage{booktabs}
\usepackage{subcaption}
\usepackage{caption}
\usepackage{float}

\usepackage[pagebackref=true,breaklinks=true,colorlinks,bookmarks=false]{hyperref}

\def\cvprPaperID{10838} %
\def\confYear{CVPR 2021}

\begin{document}
\title{Home Action Genome: Cooperative Compositional Action Understanding}

\author{%
Nishant Rai$^1$, Haofeng Chen$^1$, Jingwei Ji$^1$, Rishi Desai$^1$, Kazuki Kozuka$^2$, Shun Ishizaka$^2$, \\
Ehsan Adeli$^1$, Juan Carlos Niebles$^1$\\
$^1$Stanford University \quad $^2$Panasonic Corporation\\
{\tt\small \url{http://www.homeactiongenome.org}}
}

\maketitle

\begin{abstract}
Existing research on action recognition treats activities as monolithic events occurring in videos. Recently, the benefits of formulating actions as a combination of atomic-actions have shown promise in improving action understanding with the emergence of datasets containing such annotations, allowing us to learn representations capturing this information. However, there remains a lack of studies that extend action composition and leverage multiple viewpoints and multiple modalities of data for representation learning.
To promote research in this direction, we introduce \textit{Home Action Genome (HOMAGE)}: a multi-view action dataset with multiple modalities and view-points supplemented with hierarchical activity and atomic action labels together with dense scene composition labels. Leveraging rich multi-modal and multi-view settings, we propose Cooperative Compositional Action Understanding (CCAU), a cooperative learning framework for hierarchical action recognition that is aware of compositional action elements. CCAU shows consistent performance improvements across all modalities. Furthermore, we demonstrate the utility of co-learning compositions in few-shot action recognition by achieving 28.6\% mAP with just a single sample.
\end{abstract}

\vspace{-15pt}

\begin{figure}
\begin{center}
    \centering
    \vspace{-10pt}
    \includegraphics[width=\linewidth]{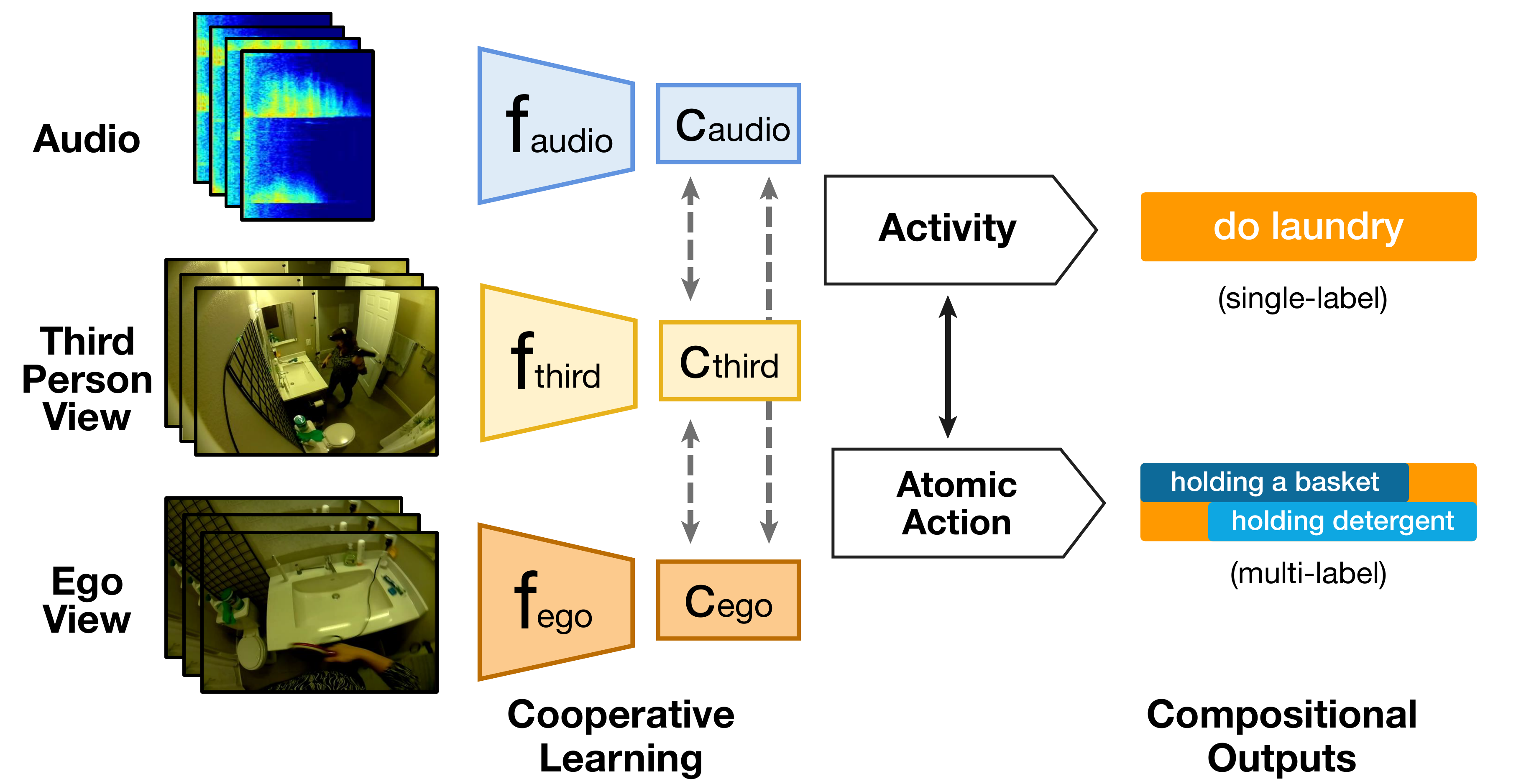}
    \vspace{-15pt}
    \captionof{figure}{\small{Given an activity instance (e.g., `do laundry') and corresponding multiple views, we compute features using modality-specific deep encoders ($f$ modules). Different modalities may capture different semantic information regarding the action. \textit{Cooperatively} training all modalities together allows us to see improved performance. We utilize training using both video-level and atomic action labels to allow both the videos and atomic actions to benefit from the \textit{compositional} interactions between the two. As discussed in the results, we see significantly improved performance when using the above components together.}}
    \label{intuition}
    \vspace{-20pt}
    \centering
\end{center}
\end{figure}

\section{Introduction}

Action understanding in videos is a critical task with various use-cases and real-world applications, from robotics \cite{rezazadegan2017action, liu2020spatiotemporal} and human-computer interaction \cite{santini2017calibme} to healthcare \cite{haque2020illuminating, lu2020vision} and elderly behavior monitoring \cite{jalal2017depth, luo2018computer}. Despite the recent success of deep learning methods for image classification, complex and holistic action or event understanding remains an elusive task.

There are several challenges associated with the task of action understanding. The inherent variability in executing complex activities poses one of the most critical difficulties in building action understating models. To understand these challenges, it is essential to understand what actions are composed of. As opposed to bounding boxes in the object detection task, actions are composed of various parts spanned in space and time. For instance, the action of ``laundry'' involves multiple entities, e.g., humans, objects, and their relationships, and is composed of a number of atomic actions. Such partonomy of actions \cite{blanke2010remember, zacks2001event, action_genome} both in space and time defines a hierarchical structure. Furthermore, to capture the variability in executing complex activities, understanding each part (e.g., body limbs, objects, or atomic actions) becomes crucial. Since actions happen in the 3D world, a holistic understanding of the world requires capturing the subtle movements or parts using multiple modalities (e.g., RGB and audio) and from multiple viewpoints.

Each of these challenges has previously been separately investigated using different datasets and advanced methods. For instance, numerous datasets were put together for generic action recognition and spatio-temporal localization in YouTube or broadcasting third-person videos, such as Kinetics \cite{kinetics700}, Charades \cite{charades}, ActivityNet \cite{activitynet}, UCF101 \cite{ucf101}. Other datasets such as EPIC Kitchens \cite{epic} were used for ego-centric action recognition. Action Genome \cite{action_genome} focused on using scene information in action recognition, while others \cite{malla2020titan} focused on hierarchical action modeling from events to low-level atomic actions. Several studies target learning from long instructional videos and release datasets \cite{chang2020procedure, miech2019howto100m, tang2019coin, ZhXuCoAAAI18} for the same, exploring the partonomy of actions in long sequences. Others also focused on observing and recognizing actions from multiple views, such as LEMMA \cite{jia2020lemma} and HumanEva \cite{humaneva}. In parallel, there have been numerous recent advances in contrastive and cooperative learning \cite{simclr, han2020self} applied to multi-modal and multi-view datasets as a self-supervised pre-training strategy to improve downstream recognition results. Despite all these advances, action understanding and generalizability of such models remains a challenging problem due to complexities brought by their complicated nature and numerous object interactions. Multi-modal approaches \cite{simonyanTwoStream14, crossLearn, cmc} have shown superior performance in tackling such issues. However, there is still a need for a benchmark that unifies all these challenges and tasks. In this paper, we release a dataset along with a novel method for hierarchical action recognition to tackle these problems.

We introduce a new benchmark for action recognition, Home Action Genome (HOMAGE), that includes multi-modal synchronized videos from multiple viewpoints along with hierarchical action and atomic-action labels. Actions in homes are challenging as we deal with long-term actions, interactions with objects, and frequent occlusions. Having multiple views and sensors to handle occlusions and scene graph information to capture object interaction allows us to tackle these complexities. In addition, synchronous videos provide implicit alignment that facilitates multi-modal training. Additionally, access to sensor information enables future research in privacy-aware recognition where we avoid audio-visual modalities. HOMAGE also provides temporal annotations of high-level activity and low-level atomic action supplemented with spatio-temporal scene-graphs. Annotations regarding interaction of objects within actions and atomic actions within high-level actions enable research in explainable video understanding, early action prediction, and long-range action recognition.

For this new benchmark, we introduce a novel method to perform simultaneous co-training with multiple modalities (RGB, audio, and annotations of scene composition) and viewpoints that enable the learning of rich video representations. Training involves a co-training strategy that leverages information from all views and modalities to build the representation space. During inference, we set up different experiments and observe improved action recognition performance even when only a single modality is used, which suggests training on HOMAGE improves performance with no need for other modalities during inference. In this paper, we explore audio-visual data (of interest to the vision community). Future sensor-fusion work can further exploit other modalities we release (e.g., for privacy-preserving studies).

HOMAGE aims to unify various aspects and challenges of action recognition, specifically targeting multi-modal and compositional perception for home actions. Moreover, the presence of a large number of modalities in our dataset encourages research in areas such as privacy-aware recognition and sensor-fusion. To summarize, our contributions are as follows: \\
\textbf{(1)} We introduce a new dataset, Home Action Genome (HOMAGE) with multiple views and modalities densely annotated with scene graphs and hierarchical activity labels (overall activity and atomic actions). \\
\textbf{(2)} We propose a novel learning framework (CCAU) that leverages multiple modalities and hierarchical action labels and improves the performance of the baselines trained on each individual modality. We demonstrate the benefits of our approach with an improvement of +6.4\% using only ego-view during inference.

\section{Related Work}

\noindent\textbf{Action Recognition in Videos.} Action recognition has continuously been an important direction for the computer vision research community. The success of 2D convolutions in image classification allowed frame-level action recognition to become a viable approach. Subsequently, two-stream networks for action recognition \cite{simonyanTwoStream14} have led to many competitive approaches, which demonstrates using multiple modalities such as optical flow helps improve performance considerably. Their work motivated other approaches that model temporal motion features together with spatial image features from videos. \cite{tran2015learning, varol2017long} demonstrated that replacing 2D convolutions with 3D convolutions leads to further performance improvements. Recent approaches such as I3D \cite{carreira2018action} inflate a 2D convolutional network into 3D to benefit from the use of pre-trained models. 3D-ResNet \cite{hara2018can} adds residual connections building a very deep 3D network leading to improved performance.

\noindent\textbf{Related Datasets.}
MSR-Action3D \cite{li2010action} provides depth map sequences containing 20 actions of interactions with game consoles. \cite{liu2017pku, ni2011rgbd, shahroudy2016ntu, sung2011human} use the Microsoft Kinect sensor to collect multi-modal action data with RGB and depth map sequences. NTU RGB+D \cite{shahroudy2016ntu} consists of RGB, depth map, infrared frames with 3D human joints annotations with 40 human subjects, and 80 distinct camera viewpoints. However, for action labels, each video in these datasets has a single video-level label and thus tough to use for action localization applications.

Other datasets \cite{activitynet, liu2017pku, kong2019mmact, jia2020lemma, action_genome} provide annotations for temporally localized actions. MMAct \cite{kong2019mmact} is a large-scale action recognition benchmark multimodal data including RGB videos, keypoints, acceleration, gyroscope, and orientation. It provides an ego-view and 4 third-person views and temporally localized actions. However, MMAct does not provide bounding box annotations for spatial localization and relationships between objects. LEMMA \cite{jia2020lemma} is a recent multi-view and multi-agent human activity recognition dataset, providing bounding box annotations on third-person views and compositional action labels annotated with predefined action templates and verbs/nouns. However, they do not provide bounding boxes of objects the subjects (human) interact with. Action Genome \cite{action_genome} is built upon the videos from Charades \cite{sigurdsson2016hollywood}, with the additional annotation of spatio-temporal scene graph labels. However, it only provides videos from a single camera view. HOMAGE aims to provide 1) multiple modalities to promote multi-modal video representation learning, 2) high-level activity labels and temporally localized atomic action labels, and 3) scene graphs that provide spatial localization cues for both the subject and the object and their relationship.

\noindent\textbf{Multi-Modal Learning.} Multiple modalities of videos are rich sources of information for both supervised \cite{simonyanTwoStream14} and self-supervised learning \cite{crossLearn, cmc, vgan}. \cite{korbar2018cooperative, cmc} introduce a contrastive learning framework to maximize the mutual information between modalities in a self-supervised manner. The method achieves state-of-the-art results on unsupervised learning benchmarks while being modality-agnostic and scalable to any number of modalities. Two stream networks for action recognition \cite{simonyanTwoStream14} have led to many competitive approaches, which demonstrate using even derivable modalities such as optical flow helps improve performance considerably. There have been approaches \cite{lt_motion, vgan, cmc, crossLearn} utilizing diverse modalities, sometimes derivable from one other, to learn better representations.

\section{Home Action Genome (HOMAGE)}
Home Action Genome (HOMAGE) is a new benchmark for action recognition that includes multi-modal synchronized video data from multiple viewpoints (ego-view,  third-person) with both high-level activity and low-level action definitions. HOMAGE focuses on actions in residential settings due to the challenges involved i.e. complexity and long duration of actions, object interactions, and frequent occlusions. HOMAGE provides multiple views and sensors to tackle these challenges. We describe the design, data collection, and data annotation process of the HOMAGE dataset in this section.

\begin{figure}[t]
    \begin{center}
    \includegraphics[width=\linewidth]{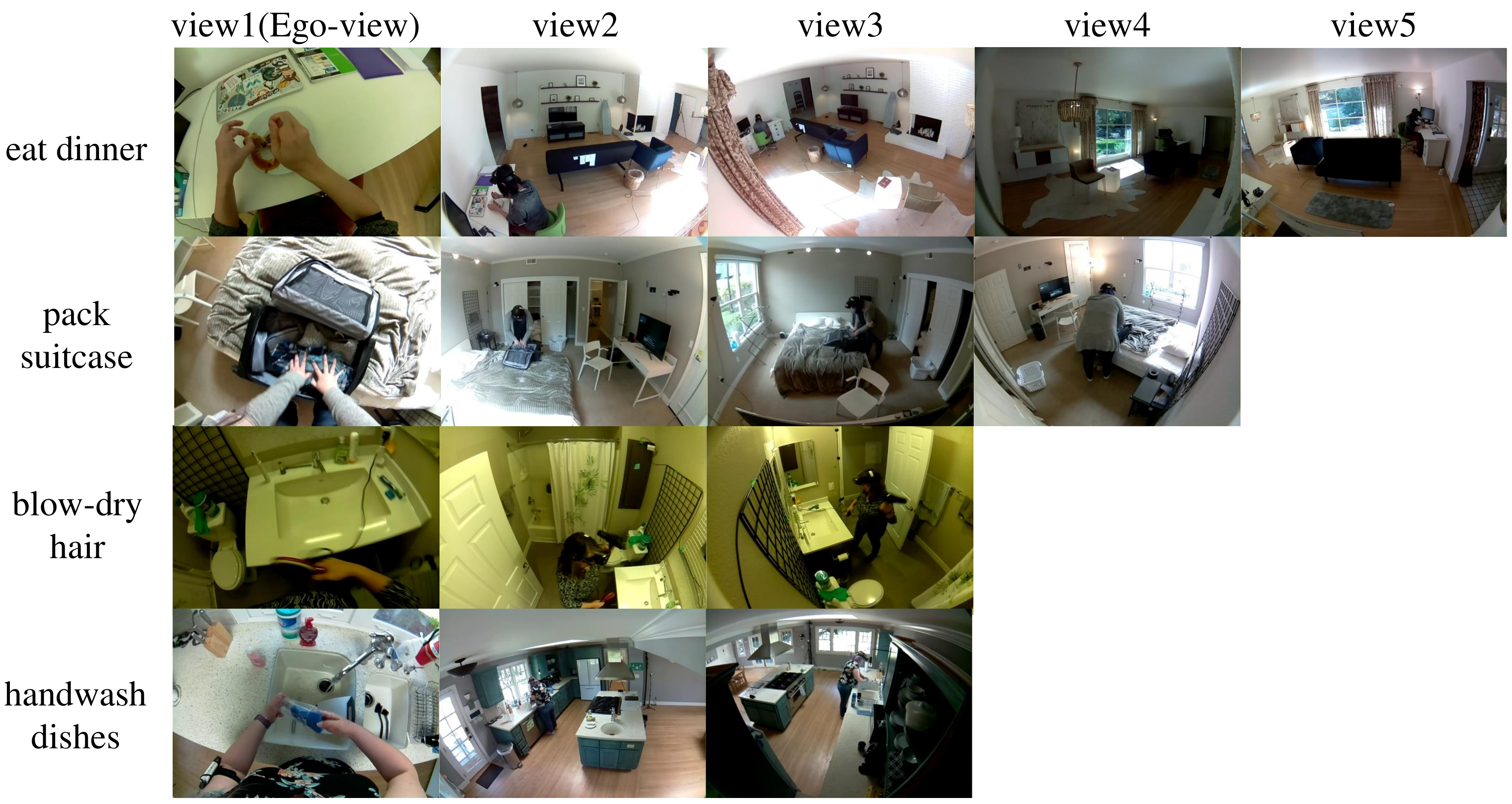}
    \vspace{-20pt}
    \caption{\small Multiple Views of Home Action Genome (HOMAGE) Dataset. Each sequence has one ego-view video as well as at least one or more synchronized third person views. \label{fig:home_action_genome}}
    \vspace{-20pt}
    \end{center}
\end{figure}

\begin{table*}[ht]
\small
\centering
\begin{tabular}{lllllllllll}
Dataset                                              & Seq   & hrs   & Modalities & Views    & HL                        & HL Classes                     & TL                        & TL Classes & TL Ins & SG                        \\\hline
RGBD-HuDaAct \cite{ni2011rgbd}      & 1.19K & 46    & 2          & 1        & \checkmark & 12                             & -                         & -          & -      & -                         \\
UCF101 \cite{ucf101}                & 13K   & 27    & 1          & 1        & \checkmark & 101                            & -                         & -          & -      & -                         \\
ActivityNet \cite{activitynet}      & 28K   & 648   & 1          & 1        & \checkmark & 200                            & -                         & -          & -      & -                         \\
Kinetics-700 \cite{kinetics700}     & 650K  & 1.79K & 1          & 1        & \checkmark & 700                            & -                         & -          & -      & -                         \\
AVA \cite{ava}                      & 430   & 108   & 1          & 1        & -                         & -                              & \checkmark & 80         & 1.58M  & -                         \\
PKU-MMD \cite{liu2017pku}           & 1.08K & 50    & 3          & 3        & -                         & -                              & \checkmark & 51         & 20K    & -                         \\
EPIC-Kitchens \cite{epic}           & -     & 55    & 1          & 1        & -                         & -                              & \checkmark & 125        & 39.6K  & -                         \\
MMAct \cite{kong2019mmact}          & 36K   & -     & 6          & 5        & -                         & -                              & \checkmark & 37         & 36.8K  & -                         \\
Action Genome \cite{action_genome} & 10K   & 82    & 1          & 1        & -                         & -                              & \checkmark & 157        & 66.5K  & \checkmark \\
Breakfast \cite{breakfast}          & -     & 77    & 1          & 1        & \checkmark & 10                             & \checkmark & 48         & -      & -                         \\
LEMMA \cite{jia2020lemma}           & 324   & 10.1  & 2          & 4        & \checkmark & 15\footnotemark & \checkmark & 863        & 11.8K  & -                         \\\hline
Ours                                                 & 1.75K & 25.4  & 12         & 2$\sim$5 & \checkmark & 75                             & \checkmark & 453        & 24.6K  & \checkmark
\end{tabular}
\vspace{-5pt}
\caption{\small Comparison between related datasets and HOMAGE. (Seq: number of synchronized sequences, Modalities: sensor modalities not including annotation data or derived data like optical flow, Views: number of synchronized viewpoints for a given sample, HL: high-level activity label (often assigned one per video), TL: temporally localized atomic action label, SG: scene graph). HOMAGE provides rich multi-modal action data, including dense annotations such as scene graphs, along with  hierarchical action labels.}
\vspace{-5pt}
\normalsize
\end{table*}

\noindent\textbf{Activities and Scenarios.}
Our goal is to build an activity recognition dataset that depicts behaviors observed in living spaces. To cover daily activities, we employed the activity taxonomy in the American Time Use Survey (ATUS) \cite{hamermesh2005data}. The ATUS taxonomy organizes activities according to two key dimensions: 1) social interactions and 2) the locations of the activities. The ATUS coding lexicon contains a large variety of daily human activities organized under 18 top-level categories such as Personal Care, Work-Related, Education, and Household activities.

Each participant was asked to perform tasks according to the instructions assigned. To make sure the behaviors are as natural as possible, we did not specify detailed procedures and time limits within the activities, and let the individual participants perform the activity freely.

\noindent\textbf{Data Collection.}
We recorded 27 participants in kitchens, bathrooms, bedrooms, living rooms, and laundry rooms in two different houses.
We used 12 sensor types: cameras (RGB), infrared (IR), microphone, RGB light, light, acceleration, gyro, human presence, magnet, air pressure, humidity, temperature. We refer to the set of data collected from a given activity with different modalities as one synchronized action sequence. Sensors were attached to several locations in the room for third-person views and to the participants' heads for ego-view. On average, there are more than 3 views per action sequence. We synchronized the sensor recordings of all views giving us synced videos which allowed for ease of use without requiring any additional post-processing.

\begin{figure}[t]
    \begin{center}
    \includegraphics[width=\linewidth]{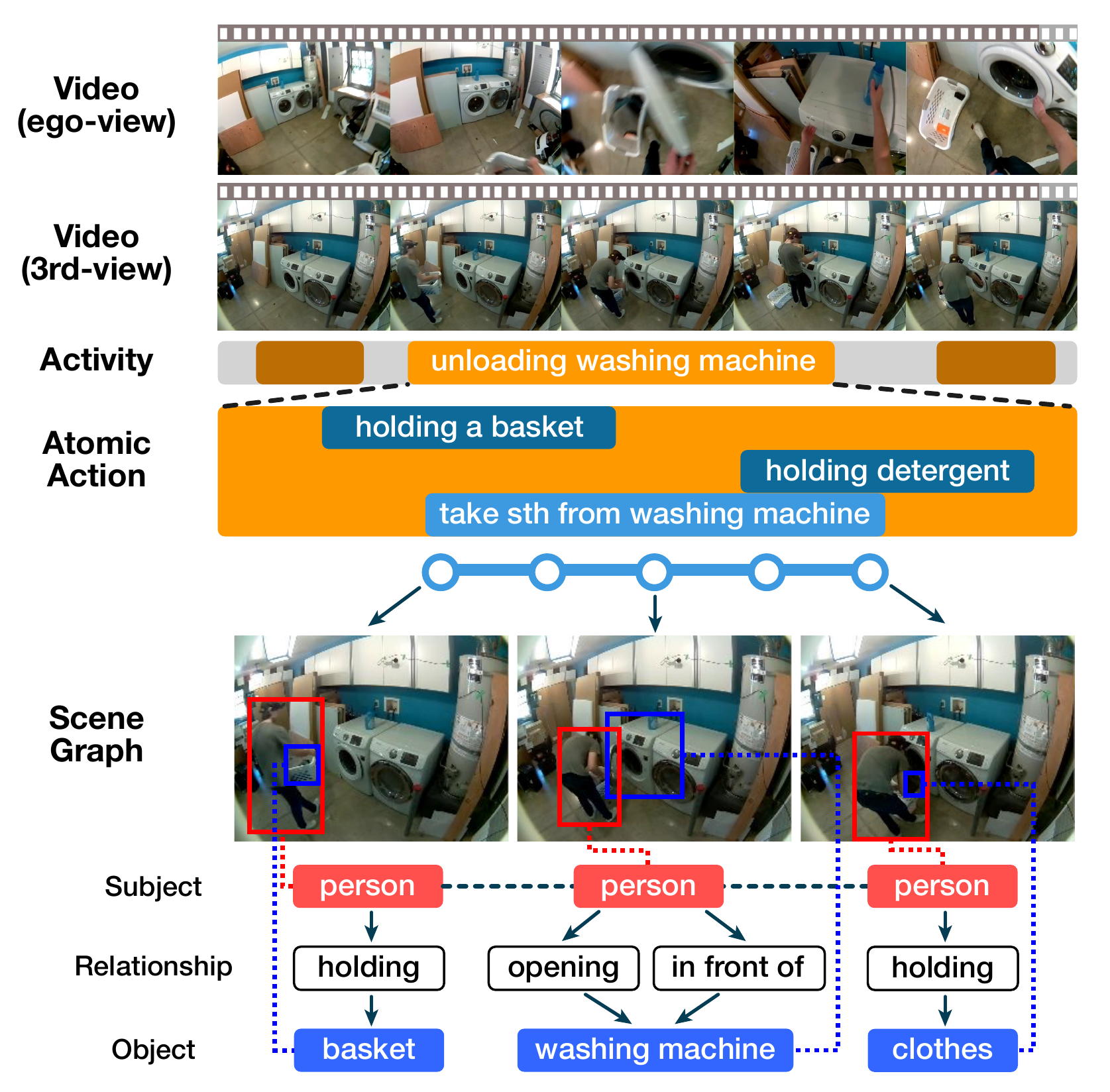}
    \vspace{-20pt}
    \caption{\small HOMAGE annotation pipeline: For every action, we uniformly sample 3 or 5 frames across the action and annotate the bounding boxes of the person performing the action along with the objects they interact with. We also annotate the pairwise relationships between the subject and the objects. \label{fig:anno_pipeline}}
    \vspace{-30pt}
\    \end{center}
\end{figure}

\noindent\textbf{Ground-truth Annotation.}
Home Action Genome is a dataset with (1) video-level activity labels, (2) temporally localized atomic activity labels, and (3) spatio-temporal scene-graph labels. Figure \ref{fig:anno_pipeline} visualizes our annotation pipeline. 
For the atomic actions, we annotated all atomic actions performed during the activities. Note that while each video can only have a single activity label, a given frame can be assigned with multiple atomic action labels when atomic actions overlap with each other. For the action graph, we annotated the person performing the action and the objects they interact with on videos from third-person views.
We uniformly sampled 3 or 5 to annotate scene graphs across the range of each atomic action interval (3 for intervals less than 3 seconds and 5 otherwise). This action-oriented dynamic sampling provides more labels where more actions occur which is very valuable for describing complex primitive actions. \cite{ji2020action} also shows this sampling scheme performs remarkably well.

\noindent\textbf{Dataset Statistics.}
We annotated 75 activities and 453 atomic actions in 1,752 synchronized sequences and 5,700 videos in total. We split the dataset into 1,388 train sequences and two test splits containing 198 and 166 sequences each. Each sequence has a high-level activity category.
We annotated atomic actions in each of these videos by providing the start and end frames and the category of the atomic action. There are 20,039 training, 2,062, and 2,468 atomic action sequences in the three splits mentioned above respectively.
For scene graphs, we annotate one third-person view video in each synchronized sequence by providing bounding boxes of the subject and the object along with the relationship between them. There are 86 object classes (excluding ``person''), and 29 relationship classes in the dataset. Overall, there are annotations of 497,534 bounding boxes and 583,481 relationships. %
\footnotetext{We here refer to the ``task classes'' in \cite{jia2020lemma}}.

The duration of atomic actions in HOMAGE is often short in time: there are about 60\% of the atomic actions under 2 seconds and 80\% under 5 seconds. For scene graphs, some of the most common objects are ``countertop,'' ``clothes,'' and ``table''; and the most common relationships include ``in front of,'' ``looking at,'' and ``holding.'' More details on the statistics are available in the supplement. 

\noindent\textbf{Relevance of Modalities.} In this paper, we only study the effect of modalities of interest to the vision community; however, HOMAGE provides rich sensor information which could be useful for privacy-aware recognition. Modalities such as angular velocity, acceleration, and geomagnetic sensors can be used to extract motion information in ego-view, and environmental sensors, e.g., temperature and humidity can capture changes in the scene before and after an activity. Thermal sensors can extract people or heat sources (e.g., extracting heat sources can be useful for recognition in places such as kitchens), and human presence and light sensors can determine the presence of people without using visual cues. Although not explored in detail in this paper, future sensor-fusion work can exploit these other modalities as well.

\section{Cooperative Compositional Action Understanding (CCAU)}

We discuss the benefits of HOMAGE and propose our approach Cooperative Compositional Action Understanding (CCAU) allowing us to exploit the rich annotations present in the dataset for improved action understanding. We discuss how CCAU employs simultaneous cooperative training with multiple modalities to improve the model's understanding of actions and the associated atomic-actions. We start by discussing a few preliminaries and proceed to discuss different components of our model. Note that ``modalities'' refer to both different camera views, as well as, modes such as images, audio, and scene graphs.

\subsection{Preliminaries}

\label{preliminaries}

A video $V$ is a sequence of $T$ frames with resolution $H \times W$ and $C$ channels, $\{\mathbf{i}_1, \mathbf{i}_2, \ldots, \mathbf{i}_{T}\}$, where $\mathbf{i}_t \in \mathbb{R}^{H \times W \times C}$. Assume $T = N * K$, where $N$ is the number of blocks and $K$ denotes the number of frames per block. We partition a video clip $V$ into $N$ disjoint blocks $V = \{\mathbf{x}_1, \mathbf{x}_2, \ldots, \mathbf{x}_{N}\}$, where $\mathbf{x}_j \in \mathbb{R}^{K \times H \times W \times C}$ and a deep encoder $f(\cdot)$ transforms each input block $x_j$ into its latent representation $z_j = f(x_j)$. An aggregation function, $g(\cdot)$ takes a sequence $\{z_1, z_2, \ldots, z_j\}$ as input and generates a context representation $c_j = g(z_1, z_2, \ldots, z_j)$. In our setup, $z_j \in \mathbb{R}^{H' \times W' \times D}$ and $c_j \in \mathbb{R}^{D}$. $D$ represents the embedding size and $H'$, $W'$ represent down-sampled resolutions as different regions in $z_j$ represent features for different spatial locations. We define $c = F(V)$, where $F(\cdot) = g(f(\cdot))$. In our experiments, $H'=4, W'=4, D=256$. The computed representations are then utilized in order to perform per-block classification to generate the necessary predictions, e.g., activity label or atomic-action label. For multiple modalities, we define $c_{m} = F_m(V_m)$, where $V_m$, $c_m$ and $F_m$ represent the video input, context feature and composite encoder for modality $m$, respectively.

\noindent\textbf{RGB Videos with Multiple Viewpoints.} An interesting aspect of HOMAGE is the presence of multiple viewpoints, specifically, a single ego-centric viewpoint and numerous third-person views. For simplicity, we treat these multiple viewpoints as two separate modalities, i.e., ego-view and third-person view. Each of these modalities has a dedicated encoder to generate clip-level features.

\noindent\textbf{Audio.} Along with having multiple camera viewpoints, we also have associated audio clips for each viewpoint. For simplicity, we only use the audio associated with the ego-centric view. For each audio clip, we generate the associated log-mel spectrogram \cite{logmel} and treat it as an image input. Following numerous other works \cite{afouras2020self, korbar2018cooperative}, we utilize a VGG19 backbone to generate a representation for the passed-in spectrogram.

\noindent\textbf{Scene Graph.} A scene graph in a given frame $G$ contains a set of objects $O = \{o_1, o_2, ...\}$ and a set of relationships $R = \{r_1, r_2, ...\}$. Each object $o_j$ contains an object ID, bounding box coordinates of the object, and object category. Each relationship $r_j$ contains the object IDs for both the subject and the object of the relationship, as well as the category of the relationship.

\subsection{Multi-Modal Cooperative Learning}

As discussed earlier, we define $c_{m} = F_m(V_m)$, where $V_m$, $c_m$ and $F_m$ represent the input, context feature, and composite encoder for modality $m$, respectively. We simultaneously train encoders for each modality while ensuring that the views improve with cooperation. Such a training regime allows us to observe improved performance during inference even when using a single modality. 

Intuitively, we expect different modalities to impart complementary information to other modalities during training. This can be similar to existing approaches such as student-teacher frameworks or knowledge distillation \cite{hinton2015distilling, kong2019mmact}. However, as we demonstrate in the experiments section, CCAU manages to learn better representations. We argue this is because the unidirectional formulation of student/teacher does not suit such setups as different modalities serve as a collective cohort of students as opposed to one of them being significantly dominant compared to others. CCAU utilizes contrastive multi-modal losses to promote cooperation between the learners.

Noise Contrastive Estimation (NCE) \cite{gutmann2010noise, mnih2013learning, oord2018representation} constructs a binary classification task where a classifier is fed with real and noisy samples with the training objective to distinguish them. We utilize a simple task of performing alignment between different modes $m, m'$. The task becomes choosing the correct in-sync instance amongst multiple noisy instances. Similar to \cite{cmc}, we use an NCE loss over our feature embeddings $c$ described in Eq.~\eqref{eq-align}. $c^{m}_{i}$ represents the feature embedding for the $m^{th}$ modality's $i^{th}$ temporal block. This effectively becomes a cross-entropy loss distinguishing one positive pair from all the negative pairs present in a video. In a batch setting with multiple video clips, it is possible to have more inter-clip negative pairs.  The objective function for a single pair of modalities will hence be:
{\small \begin{equation}
    \label{eq-align}
    \mathcal{L}^{m, m'}_{align} = - \sum_{i} \bigg( \text{log} \frac{\text{exp}({c}^{m}_{i} \cdot c^{m'}_{i})}{\sum_{j} \text{exp}({c}^{m}_{i} \cdot c^{m'}_{j})} \bigg).
\end{equation}}
To extend this to multiple views, we utilize the same objective for all pairs and simultaneously optimize: 
$$\mathcal{L}_{align} = \sum_{m, m'} \mathcal{L}^{m, m'}_{align}.$$ 

Self-supervised attention \cite{multisensory2018} has been shown to be useful to auto-learn associations between different modalities. We model attention by predicting importance weights over the grid. We predict $H' \times W'$ values $\alpha_{i, j}$ representing weights of each feature corresponding to spatial location $(i, j)$. Given feature $c$ of shape $D \times H' \times W'$, we extract $c_{agg}$ from it as given in Eq.~\eqref{attention}. Where $\tau$ refers to the temperature. Further details are provided in the appendix.

\begin{equation}
    \label{attention}
    c_{agg} = \sum_{i, j} p_{i, j} \cdot c_{i, j} \;\; \text{,} \;\; p_{i, j} = \frac{\text{exp}(\alpha_{i, j} \mathbin{/} \tau)}{\sum_{a, b} \text{exp}(\alpha_{a, b} \mathbin{/} \tau)}
\end{equation}

\subsection{Compositional Action Recognition}

\label{compose}

In addition to the multi-modal nature of HOMAGE, another one of its differentiating factors is having fine-grained atomic-action labels along with video-level action labels. The compositional nature of atomic-actions is useful in determining both the overall activity as well as learning relationships between atomic-actions and high-level actions.

We leverage the compositionality of atomic-actions and activities in CCAU by simultaneously utilizing both activity and atomic action level labels in our learning task. The intuition being our model will be able to learn the composition and relationships between atomic-actions and activities improving its understanding. We utilize the contextual features $c$ in order to predict class labels for both video and atomic-action classes. The video action prediction task is a standard one-hot classification task, while we formulate the atomic-action prediction task as multi-target classification. We represent their corresponding losses as $\mathcal{L}_{video} = \mathcal{L}_{v}$ and $\mathcal{L}_{atomic} = \mathcal{L}_{a}$. The overall compositional loss is represented by $\mathcal{L}_{composition} = \mathcal{L}_{c}$.

We explore two variants to define $\mathcal{L}_{c}$. The first involves manually chosen hyper-parameters modulating each component, i.e., $\mathcal{L}_{c} = \mathcal{L}_{v} + \lambda \mathcal{L}_{a}$. The second automatically learns the appropriate multi-task weights \cite{kendall2018multi}. The numbers reported in the paper represent use the first approach with $\lambda = 10$. For details refer to the appendix.

\section{Experiments}

We discussed the rich annotations in Home Action Genome (HOMAGE) that allows us to explore multiple aspects previously not possible due to the lack of such datasets. CCAU utilizes cooperative and compositional learning to learn improved representations for action understanding. Co-training with other modalities such as audio imparts additional structure and knowledge to individual modalities, also leading to improved single-view performance. We design and discuss multiple quantitative experiments to verify the validity of our claims. We also conduct qualitative experiments to gain deeper insights into our approach. In this section, we briefly go over our experiment framework. Additional details are provided in the appendix.

\subsection{Implementation Details}
Following our design discussed earlier to allow inference using individual modalities, we use separate encoders for each. We use different designs as mentioned in Section \ref{preliminaries}.

\noindent\textbf{Images.} In all of our experiments, we treat ego-view as one modality and all third-person view videos as another. We resize each input frame to the size of 128x128. We employ a 3D-ResNet similar to \cite{hara2018can} as the encoder $f(\cdot)$. Following \cite{dpc}, we only expand the convolutional kernels present in the last two residual blocks to be 3D ones and use 3D-ResNet18 for our experiments, denoted as ResNet18. A weak aggregation function $g(\cdot)$ is used to learn a strong encoder $f(\cdot)$. Specifically, we use a one-layer Convolutional Gated Recurrent Unit (ConvGRU) with kernel size (1, 1) as $g(\cdot)$. The weights are shared amongst all spatial positions in the feature map. This design allows the aggregation function to propagate features in the temporal axis. 

We use a dropout \cite{dropout} with $p = 0.1$ to compute the hidden state at each time step. A shallow two-layer perceptron is used as the predictive function $\phi(\cdot)$. Recall $z_j' = Pool(z_j)$ where $z'_j \in R_D$. We utilize stacked max pool layers as $Pool(\cdot)$. To construct blocks to pass to the network, we uniformly choose one out of every 3 frames. Then, they are grouped into 8 blocks containing 5 frames each. Since the videos are usually 30fps, each block roughly covers 0.5 seconds and 8 blocks sums to about 4 seconds worth of action. Given the 256D final representations, we pass this through fully connected layers to compute the final classification where we use a dropout of $p=0.5$.

\noindent\textbf{Audio.} To process audio clips, we convert audio to MP3 format, compute log-mel spectrograms \cite{logmel}, and pass it through a VGG19-like convolutional architecture. We sample fixed intervals of the spectrogram image to represent the action clip. Similar to the image encoder, we have fully connected layers to perform classification.

\vspace{-1pt}

\subsection{Quantitative Results}

In this section, we analyze various aspects of our proposed model. To objectively evaluate model performance, classification accuracy is utilized as a proxy for learned representation quality. Evaluation is performed on two different splits of HOMAGE. Although models have access to other modalities during training, this is not the case during inference. Therefore, evaluation only involves inference using individual modalities. However, we see an improvement despite this constraint due to co-training. We also study the improvement imparted through compositional learning with both high-level action and atomic-action labels.

\vspace{-5pt}
\subsubsection{Comparisons with Baselines}

\begin{table}[t]
\centering
\renewcommand{\arraystretch}{1.0}
\renewcommand{\tabcolsep}{2mm}
\small
\begin{tabular}{c|c|c|c}
    \toprule
    Method & Audio & Ego & $3rd$ Person \\
    \midrule
    Single Modality    & 28.5 & 31.3 & 21.8 \\
    Cooperative Ours  & \textbf{33.3} & \textbf{37.7} & \textbf{24.7} \\
    \midrule
    Static KD       & 28.5 & 32.3 & 21.8 \\
    Cooperative KD  & 32.1 & 32.1 & 23.5 \\
    \bottomrule
\end{tabular}
\normalsize
\vspace{-5pt}
\captionof{table}{{\small Video classification accuracy. \textit{Cooperative Ours} outperforms the baselines. \textit{Cooperative KD} performs better than its counterparts, further validating benefits of cooperative learning.}}
\vspace{-10pt}
\label{table:baselines}
\end{table}

In this section, we investigate the effectiveness of cooperative multi-modal learning for action understanding. We study the impact of cooperative learning and compare the performance to knowledge distillation approaches.

\noindent\textbf{Impact of Cooperation.} Our co-operative training approach hinges on the assumption that multi-modal information helps in improving overall representation quality. To verify our hypothesis, we study the performance of CCAU compared along with a few other comparable approaches. {(1) \textit{Single Modality Training (SM)}} - Training of modalities independently {(2) \textit{Cooperative Ours Training (CT)}} - Co-Training of all modalities and individual inference. Table ~\ref{table:baselines} summarizes our results demonstrating a consistent improvement in performance across modalities.

\noindent\textbf{Comparison with Knowledge Distillation.} Given the potential applicability of student-teacher approaches in this setting, we also study their performance compared to our approach. We study two variants. {(1) \textit{Static Knowledge Distillation (SKD)}} - We transfer knowledge from other trained modalities into the ego-view encoder. {(2) \textit{Cooperative Knowledge Distillation (CKD)}} - To isolate the effect of cooperation leading to improved performance, we also propose a cooperative version of knowledge distillation that allows all modalities to simultaneously improve (details in the appendix). Table ~\ref{table:baselines} summarizes our results demonstrating the performance difference between these approaches. We notice a performance improvement when utilizing cooperative KD compared to the static variant. CT outperforms CKD even though both allow cooperation, due to the incorrect student-teacher hierarchy even with a symmetric knowledge distillation setup. CT allows cooperation in a softer manner without an implicit assumption of hierarchy.

\subsubsection{Impact of Additional Modalities}

We saw the benefits of Cooperative Training in the previous section and established the performance improvements accompanying training with multiple modalities. In this section, we look at the implications modalities have on performance by studying the impact of training with multiple modalities. We consider 1) \textit{Training each modality separately}; 2) \textit{Joint training of multi-camera views}, i.e., Ego and $3rd$ Person RGB video clips, and 3) \textit{Joint training of multi-camera views with ego-centric audio clips}.

\noindent\textbf{Activity Classification.} Table \ref{table:modalities-video_level} summarizes the results of our approach trained with different modalities. Compared with training with single views individually, co-training with the two video views and video + audio consistently improves the performance together with more modalities.

\noindent\textbf{Atomic Action Classification.} We also investigate the impact of cooperative training on multi-target classification for atomic actions. Table ~\ref{table:modes+atomic} summarizes our results. The Mean Average Precision scores for each modality are reported.

\begin{table}[tb]
    \centering
    \renewcommand{\arraystretch}{1.0}
    \renewcommand{\tabcolsep}{2mm}
    \centering
    \small
    \begin{tabular}{c|c|c|c} 
        \toprule
        Method & Audio & Ego & $3rd$ \\
        \midrule    
         Single Modality          & 28.5 & 31.3 & 21.8 \\
         Coop - Ego + $3rd$       & -    & 35.1 & 23.5 \\
         Coop - Ego + $3rd$ + Aud & \textbf{33.3} & \textbf{37.7} & \textbf{24.7} \\
      \bottomrule
    \end{tabular}
    \normalsize
    \vspace{-10pt}
    \captionof{table}{ \small Co-training encoders with different modalities on activity classification. We see a distinct performance improvement across modalities as we co-train with increasing number of modes, possibly due to the presence of rich complementary information.}
    \label{table:modalities-video_level}
\end{table}

\begin{table}[tb]
    \centering
    \renewcommand{\arraystretch}{1.0}
    \renewcommand{\tabcolsep}{2mm}
    \centering
    \small
        \begin{tabular}{c|c|c|c}
        \toprule
        Method & Audio & Ego & $3rd$ Person \\
        \midrule
         Single Modality  & 7.0  & 20.5 & 11.7 \\
         Cooperative & \textbf{13.2} & \textbf{28.5} & \textbf{15.3} \\
        \bottomrule
        \end{tabular}
        \normalsize
        \vspace{-5pt}
        \captionof{table}{\small Effect of co-training encoders with different modalities on atomic action classification. The numbers reported are support weighted mAP scores.}
        \label{table:modes+atomic}
        \vspace{-10pt}
\end{table}

\begin{table}[tb]
    \centering
    \renewcommand{\arraystretch}{1.0}
    \renewcommand{\tabcolsep}{2mm}
    \centering
    \small
    \begin{tabular}{c|c|c}
      \toprule
        Method & Ego & $3rd$ Person \\
        \midrule
         Cooperative                & 32.5 & 19.1 \\
         Cooperative with Attention & \textbf{34.8} & \textbf{20.8} \\
       \bottomrule
    \end{tabular}
    \normalsize
    \vspace{-5pt}
    \captionof{table}{\small Effect of co-training encoders using the proposed attention module. We see a consistent performance improvement across both modalities. The $3rd$ person mode benefits as attention allows potential localization of the region of interest - despite the lack of dense associations between the ego and $3rd$ person view.}
    \label{table:attention-video_level}
    \vspace{-15pt}
\end{table}

\begin{table*}[tb]
    \centering
    \renewcommand{\arraystretch}{1.0}
    \renewcommand{\tabcolsep}{2mm}
    \centering
    \small
    \begin{tabular}{c|c|c|c|c|c|c} 
      \toprule
        \multirow{2}{*}{Method} & \multicolumn{3}{c|}{Acc} & \multicolumn{3}{c}{mAP} \\ & {Audio} & Ego & $3rd$ Person & {Audio} & Ego & $3rd$ Person \\
        \midrule
         Cooperative - Activity & 28.3	& 31.1 & {17.0} & - & - & - \\
         Cooperative - Atomic Actions & - & - & - & 5.9 & 18.5 & 9.5 \\
         Compositional          & 23.5 & 32.1 & 16.2 & 16.4 & 26.3 & {12.2} \\
         Cooperative Compositional & \textbf{29.3} & \textbf{34.9} & \textbf{19.2} & \textbf{21.7} & \textbf{29.3} & \textbf{13.8} \\
      \bottomrule
    \end{tabular}
    \vspace{-5pt}
    \captionof{table}{\small Effect of co-training encoders with images and audio on activity classification. We see a distinct performance improvement compared to the Ego, $3rd$ Person Co-Training case; due to the rich complementary information present in audio encoders. Missing numbers denote the model was not trained for the associated subtask. Results are averaged over the two test splits.}
    \vspace{-10pt}
    \label{table:composition}
    \normalsize
\end{table*}

\begin{table}[tb]
\centering
\renewcommand{\arraystretch}{1.0}
\renewcommand{\tabcolsep}{2mm}
\small
\begin{tabular}{c|c|c|c|c}
    \toprule
    \multirow{2}{*}{Method - Ego} & \multicolumn{4}{c}{Atomic Action - mAP}\\
    & \multicolumn{1}{c|}{1 shot} & 5 shot & 10 shot & 20 shot \\
    \midrule
    Single Modality & 22.4 & 35.3 & 38.6 & 40.6 \\
    CCAU            & \textbf{28.6} & \textbf{36.9} & \textbf{39.4} & \textbf{49.4} \\
    \bottomrule
\end{tabular}
\normalsize
\vspace{-5pt}
\captionof{table}{\small{Compositional learning with few shot learning. With compositional action understanding, CCAU demonstrates much better generalizability than other baseline, showing the potential of co-learning with compositional labels in improving action understanding. Results are averaged over the two testing splits.}}
\vspace{-10pt}
\label{table:few-shot}
\end{table}

\subsubsection{Cooperative Compositional Learning}

We analyze the role of both our proposed soft attention module and CCAU's compositional learning framework.

\noindent\textbf{Impact of Co-training with Attention.} Table \ref{table:attention-video_level} summarizes the results of the cross-modality co-training experiment with and without attention module. With attention, the model yields better accuracy on the video modalities compared with its counterpart. The model can implicitly learn localization and correspondence between views to form representations with view-invariant information.

\noindent\textbf{Impact of Compositional Learning.} Our compositional learning framework hinges on the assumption that simultaneously learning both activity labels and atomic action labels leads to improved performance. To verify this hypothesis, we compare different variants such as (1) train with activity labels, (2) train with atomic-action labels, (3) train with both activity and atomic actions without cooperation and (4) CCAU - cooperatively train with both video and atomic actions. In Table ~\ref{table:composition}, we see a consistent improvement across both activity and atomic-action performance.

\subsubsection{Few-Shot Compositional Action Learning}

We have discussed the benefits of our cooperative and compositional approach. Intuitively, predicting activities should be easier if we have an idea of the atomic-actions composing the higher-order action. We now showcase the ability and potential of CCAU to generalize to rare actions.

\noindent\textbf{Setup.} In our few-shot action recognition experiments, we split the 75 action classes into a base set of 60 classes and a novel set of 15 classes. We use CCAU as our feature extractor. Note that we do not finetune the backbone. Next, we train each model with only k examples from each novel class, where k = 1, 5, 10, 20. Finally, we evaluate the trained models on all examples of novel classes in the validation set.

\noindent\textbf{Results.} We report few-shot experiment performance in Table ~\ref{table:few-shot}. CCAU improves the single modality baseline on all 1, 5, 10, 20-shot experiments. Furthermore, CCAU shows a +6.2\% 1-shot and +8.8\% 20-shot mAP improvement.

\subsection{Qualitative Results}

One of the motivating factors behind CCAU was the benefits of co-training different encoders together to gain higher-order perspectives provided through different modalities. We observe the learned structure across modalities results in the emergence of higher-order semantics without additional supervision, e.g., sensible class relationships and good feature representations. Jointly training with modalities gives rise to better representations and byproducts such as localization of visual regions of interest.

\begin{figure}
\begin{center}
    \centering
    \includegraphics[width=\linewidth]{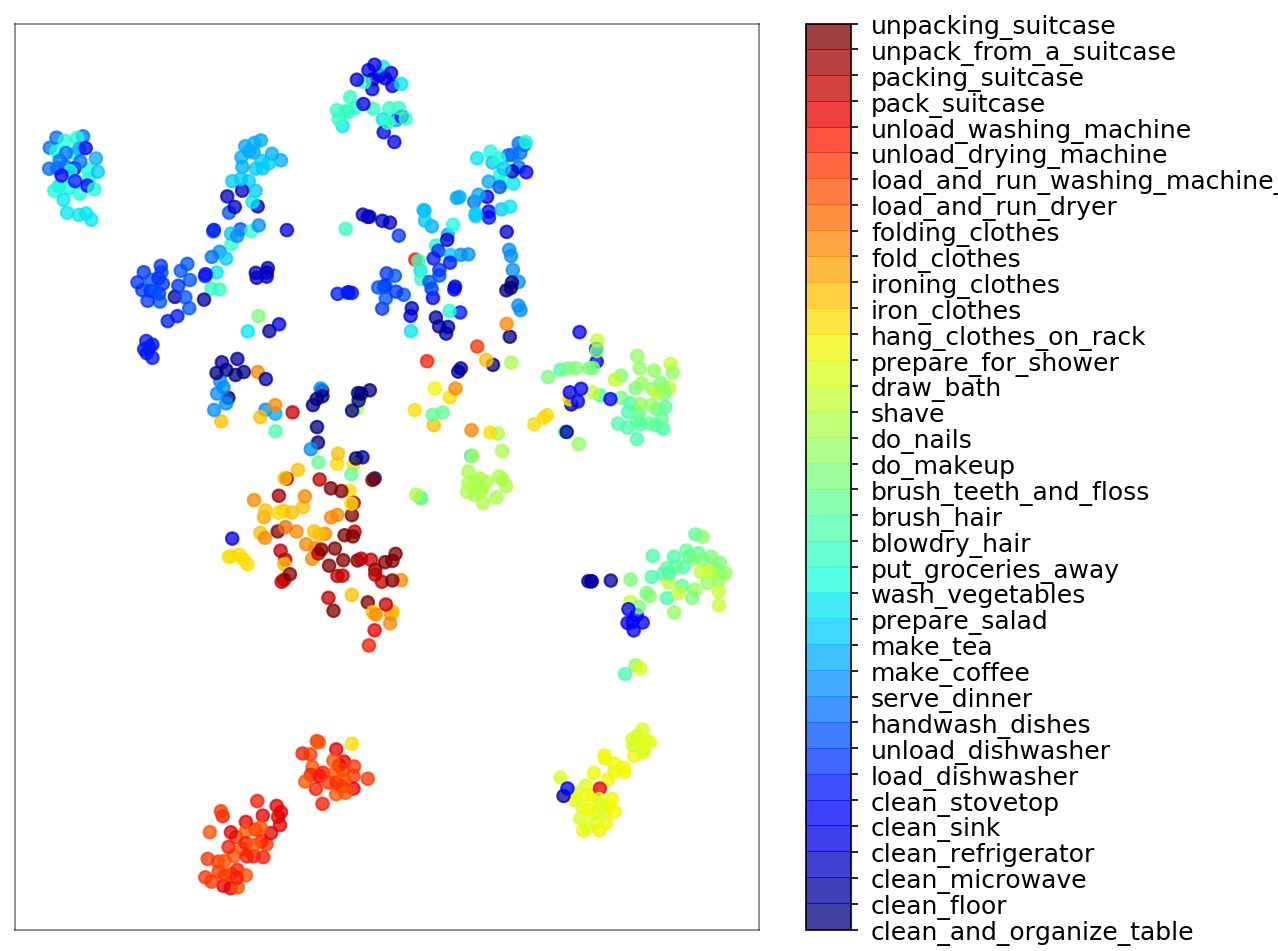}
    \vspace{-20pt}
    \captionof{figure}{\small{t-SNE visualization of Ego-View features from \textit{CCAU} trained with ego, $3rd$ and audio modalities. The color mapping represents the relationships between the action classes, e.g., Red: Clothes; Green: Grooming; Blue: Kitchen. CCAU is able to learn meaningful clusters by utilizing compositional information.}}
    \label{tsne}
    \vspace{-10mm}
    \centering
\end{center}
\end{figure}

\noindent\textbf{t-SNE Visualization.} We explore t-SNE visualizations of our learned representations. For clarity, only a few action classes are displayed. We loosely order the action classes according to their relationships; classes having similar colors are semantically similar. Fig.~\ref{tsne} summarizes our results.

\noindent\textbf{Multi-Modal Localization.} A by-product of learning attention using contrastive losses is the ability to localize potential points of interest in images (details in the appendix).

\section{Conclusion}

We introduced Home Action Genome (HOMAGE), a human action recognition benchmark with multiple modalities and viewpoints with hierarchical activity and atomic action labels. We also proposed CCAU, a cooperative and compositional learning method to leverage information across multiple modalities along with action compositions in HOMAGE for better representation learning. Due to the nature of cooperative learning, CCAU allows inference on individual modalities where no privileged information and other modalities are available. We demonstrated the benefits of learning atomic-actions compositions leading to significantly improved results in a few-shot learning setting.

With rich multi-modal data and compositional annotations, HOMAGE facilitates research in subfields such as multi--modal action recognition and localization, explainable action understanding, and reasoning with spatio-temporal scene graphs. We hope HOMAGE promotes research in multi-modal cooperative learning and action understanding using compositions for richer feature representations in human action recognition as well as raises interest in generalizable video understanding.

\noindent\textbf{Acknowledgement.}
This work has been supported by Panasonic. This article solely reflects the opinions and conclusions of its authors and not Panasonic or any entity associated with Panasonic.

{\small
\bibliographystyle{ieeetr}
\bibliography{cvpr}
}

\clearpage

\renewcommand\thefigure{A.\arabic{figure}}    
\setcounter{figure}{0}   
\renewcommand\thetable{A.\arabic{table}}    
\setcounter{table}{0}   

\section*{Appendix A - Dataset}
\label{sec:dataset}

\subsubsection*{Sensors and Modalities}

\begin{figure}[b]
\begin{center}
    \centering
    \includegraphics[width=\linewidth]{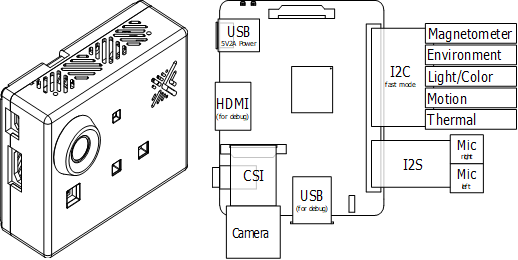}
    \captionof{figure}{Multi-modal sensor kit used in data collection.}
    \label{fig:sensor}
    \centering
\end{center}
\end{figure}

\begin{figure}[t]
\begin{center}
    \centering
    \includegraphics[width=\linewidth]{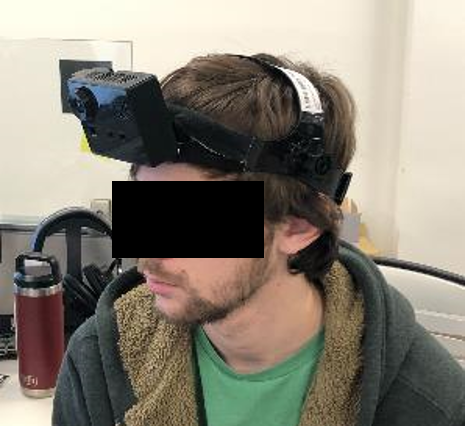}\vspace{-10pt}
    \captionof{figure}{The sensor, mounted on the participant's head.}
    \label{fig:sensor_photo}
    \centering
\end{center}
\end{figure}

We build multi-modal sensor kits for data collection as shown in Figure \ref{fig:sensor}. This kit assists the creation of the multi-modal dataset by dramatically simplifying the data collection process through simple recording and timing synchronization.  The data from all viewpoints are collected by these sensor-kits. Figure \ref{fig:sensor_photo} shows the photo of the multi-modal sensor mounted on the head of a subject participant.

The audio and video data from the sensor is saved to a video file, and the sensor data is saved in the same file as additional tracks. By using lossless codecs like the Free Lossless Audio Codec (FLAC) or WavPack, we can save the sensor data with high fidelity. Both codecs support multi-channel audio in 8-32 bit integer format at frequencies as low as 1Hz. Sensor data is acquired over I2C with constant timing adjustments to maintain synchronization with audio and video.

HOMAGE contains 12 modalities with multiple viewpoints. Specifically, the infrared data is obtained by the Grid-EYE 8x8 pixel infrared array sensor. The RGB light data is obtained by a photodiode array sensor that provides an RGB spectral response with IR blocking filter. The sensor kit also includes an ambient light sensor that combines a broadband photodiode and an infrared-responding photodiode on a single CMOS-integrated circuit to provide ambient light data. The human presence sensor is a 4-channel nondispersive infrared (NDIR) sensor. The magnetic field data is acquired from a magnetometer in the sensor kit.

\subsubsection*{Data Collection}

We collect human action data from different viewpoints using our multimodal sensor kits. We provide additional details in Table ~\ref{table:sensors}. Specifically, we synchronize the data from different modalities by using the following scheme. 
(1) The participants were instructed to start the activity displayed on the screen after they heard the start tone.
(2) The content of the participants was specified by activity unit (e.g. make bed). We do not specify a detailed sequence of atomic actions.
(3) We sounded the end tone when the participant's activity is finished. We synchronized the data of multiple sensor-kits using the signal of start/end tone.

It is worth noting that in order to measure natural activities in a situation where we are in control of the collecting location and the objects, we did not give instructions to the actors as much as possible. We do not provide any sequence of actions or objects to touch. Furthermore, to match the activity labels like ``make bed'', the activity instructions were presented in text by display, and the actors did what they could imagine with the activity.

\begin{table}[t]
\centering
\renewcommand{\arraystretch}{1.0}
\renewcommand{\tabcolsep}{2mm}
\footnotesize
\caption{List of sensors in our multi-modal sensor}
\vspace{-10pt}
\label{table:sensors}
\begin{tabular}{c|c|c}
    \toprule
    \multirow{2}{*}{Sensor} & \multicolumn{2}{c}{Sensor information}\\
    & \multicolumn{1}{c|}{Model no.} & rate\\
    \midrule
    Video & OmniVision OV5647 & 30fps \\
    I2S Digital Microphones & SPH0645LM4H & 48KHz \\
    GridEYE Thermal Imager & AMG8833 & 10Hz \\
    Human Presence (PIR) & AK9753AE & 2Hz \\
    Ambient Light Intensity & TSL2591 & 2Hz \\
    Ambient Color & ISL29125 & 10Hz \\
    CO2/Humidity/Pressure/Temp. & BME680 & 5Hz \\
    Magnetometer & MLX90393 & 10 Hz \\
    \bottomrule
\end{tabular}
\normalsize

\end{table}

\subsubsection*{Data Synchronization}

When storing video, audio, and sensor data together, each data stream is stored in a container by multiplexing the streams. We use H264 for the video stream, and FLAC (Free Lossless Audio Codec) for audio and sensor data.

To synchronize the sensor data, A 60Hz, fixed-length time-division multiplexing scheduler is used to query the sensors over the inter-integrated circuit (I2C) bus. The scheduler monitors the drift between expected and actual query times and adjusts its timing on the fly to achieve sub-millisecond accuracy on average. Sensor data are timestamped and passed to the main thread and encoded into its respective track immediately to guarantee synchronization.

\begin{figure}[t]
\begin{center}
    \centering
    \includegraphics[width=\linewidth]{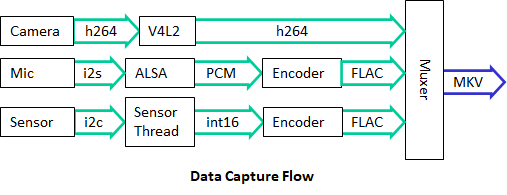}
    \captionof{figure}{The flow chart of data collection.}
    \label{fig:data_flow}
    \centering
\end{center}
\end{figure}

\subsubsection*{Data Statistics}

\begin{figure}[t]
    \begin{center}
    \includegraphics[width=\linewidth,trim=10 10 10 10 ,clip]{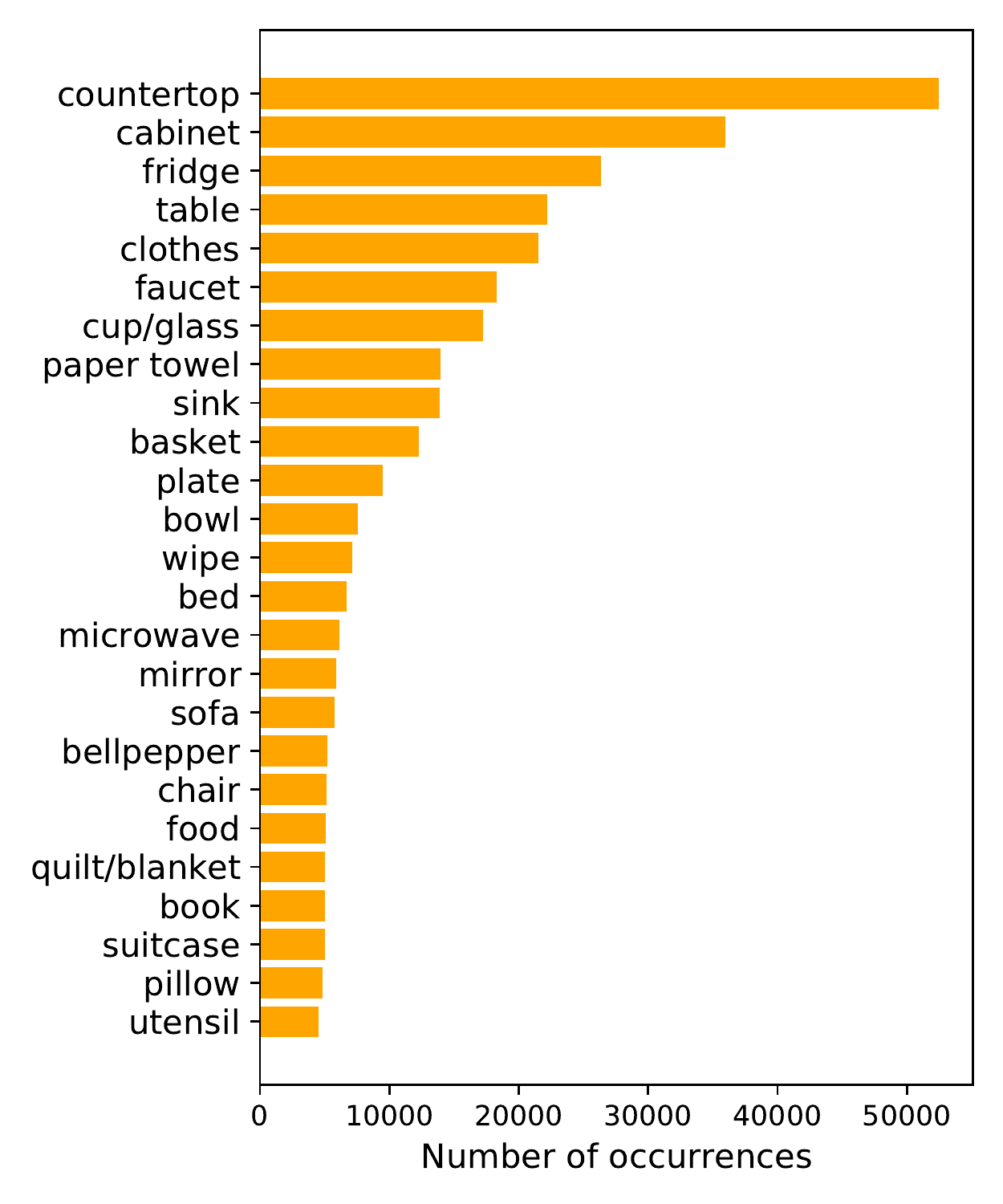}
    \caption{Distribution of object classes (top 25). \label{fig:obj_dist}}
    \end{center}
\end{figure}

\begin{figure}[t]
    \begin{center}
    \includegraphics[width=\linewidth,trim=10 10 10 10 ,clip]{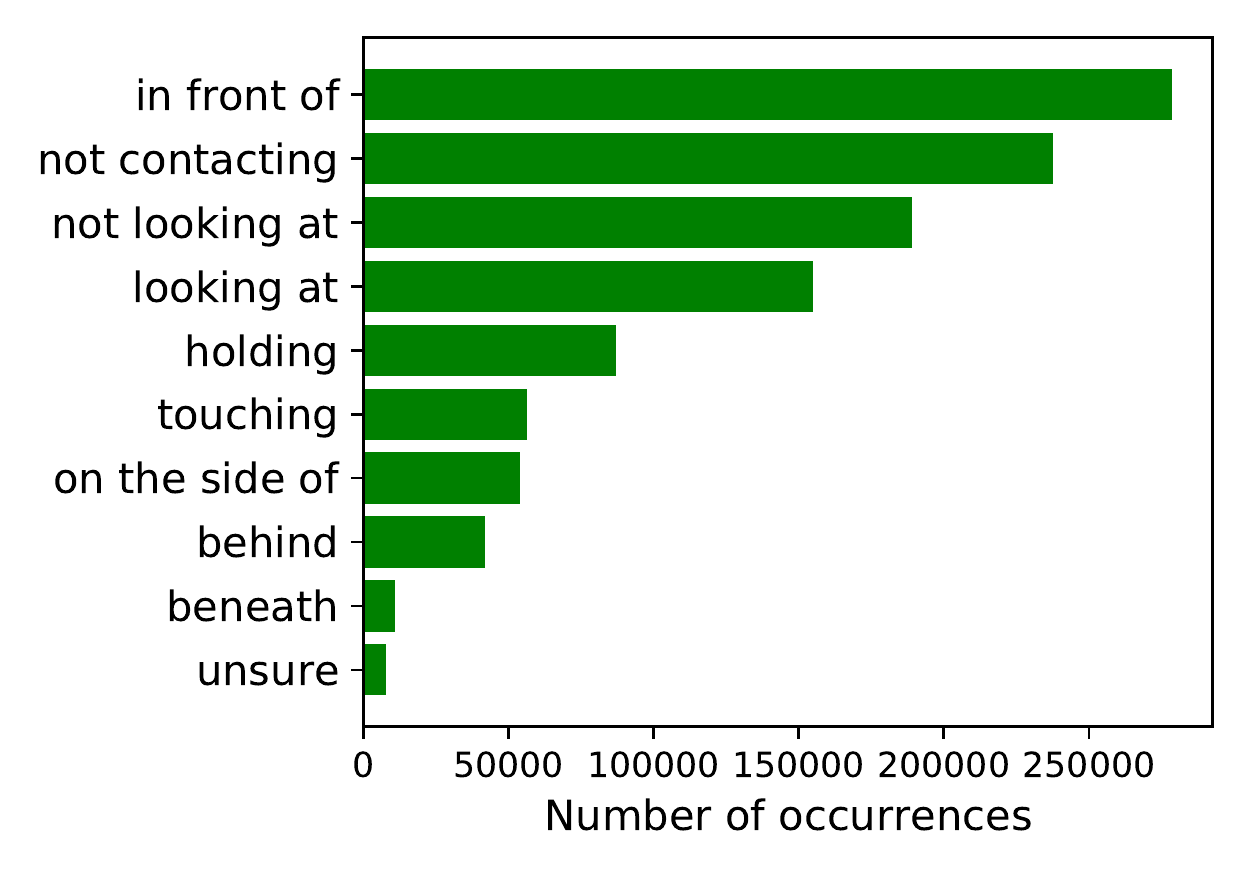}
    \caption{Distribution of relationship classes (top 10). \label{fig:rel_dist}}
    \end{center}
\end{figure}

\begin{figure*}[t]
    \begin{center}
    \includegraphics[width=0.65\linewidth,trim=20 20 110 20 ,clip]{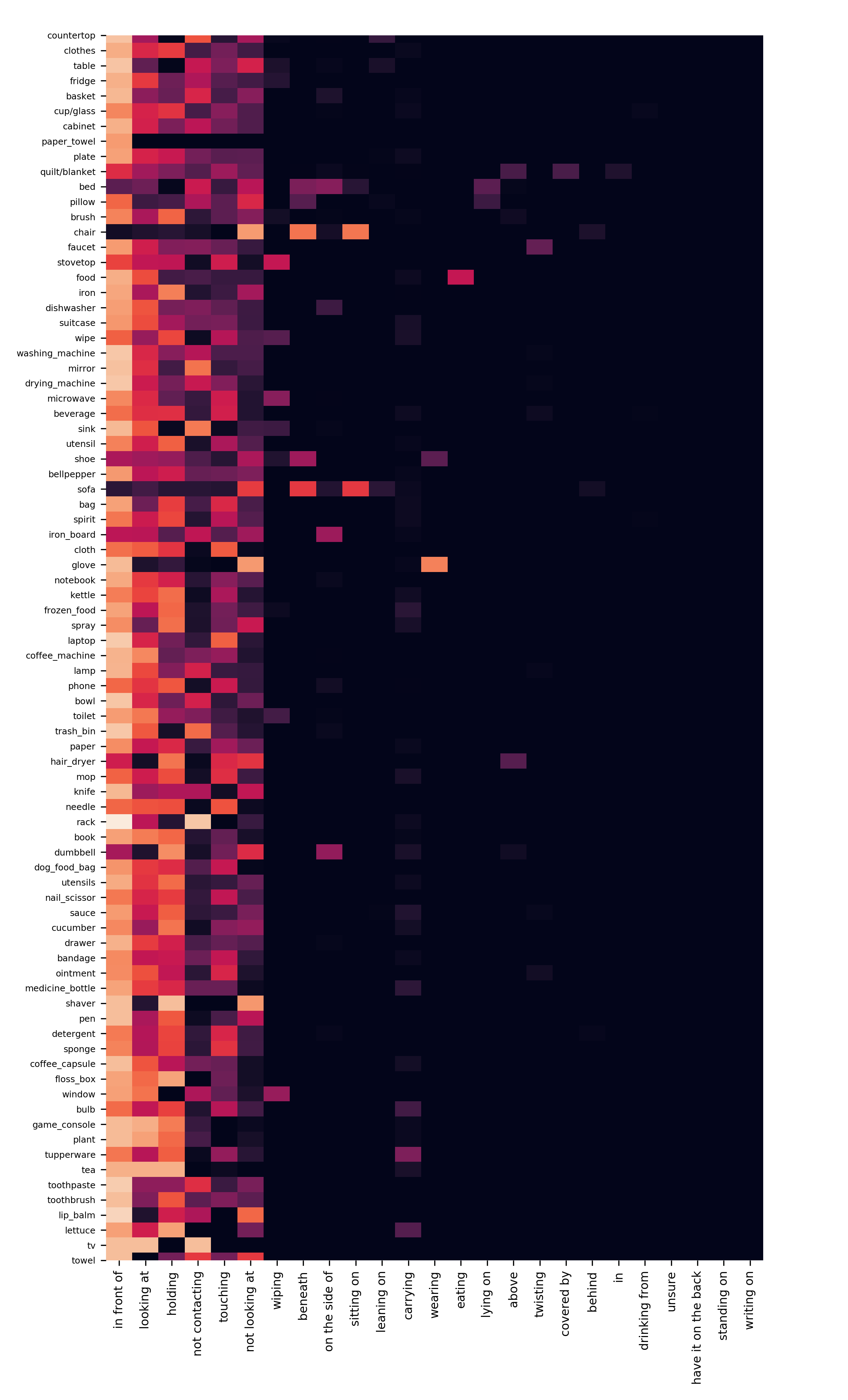}
    \caption{The co-occurrence statistics for objects and relationships in Home Action Genome. \label{fig:matrix_obj_rel}}
    \end{center}
\end{figure*}

\begin{figure}[t]
    \begin{center}
    \includegraphics[width=\linewidth]{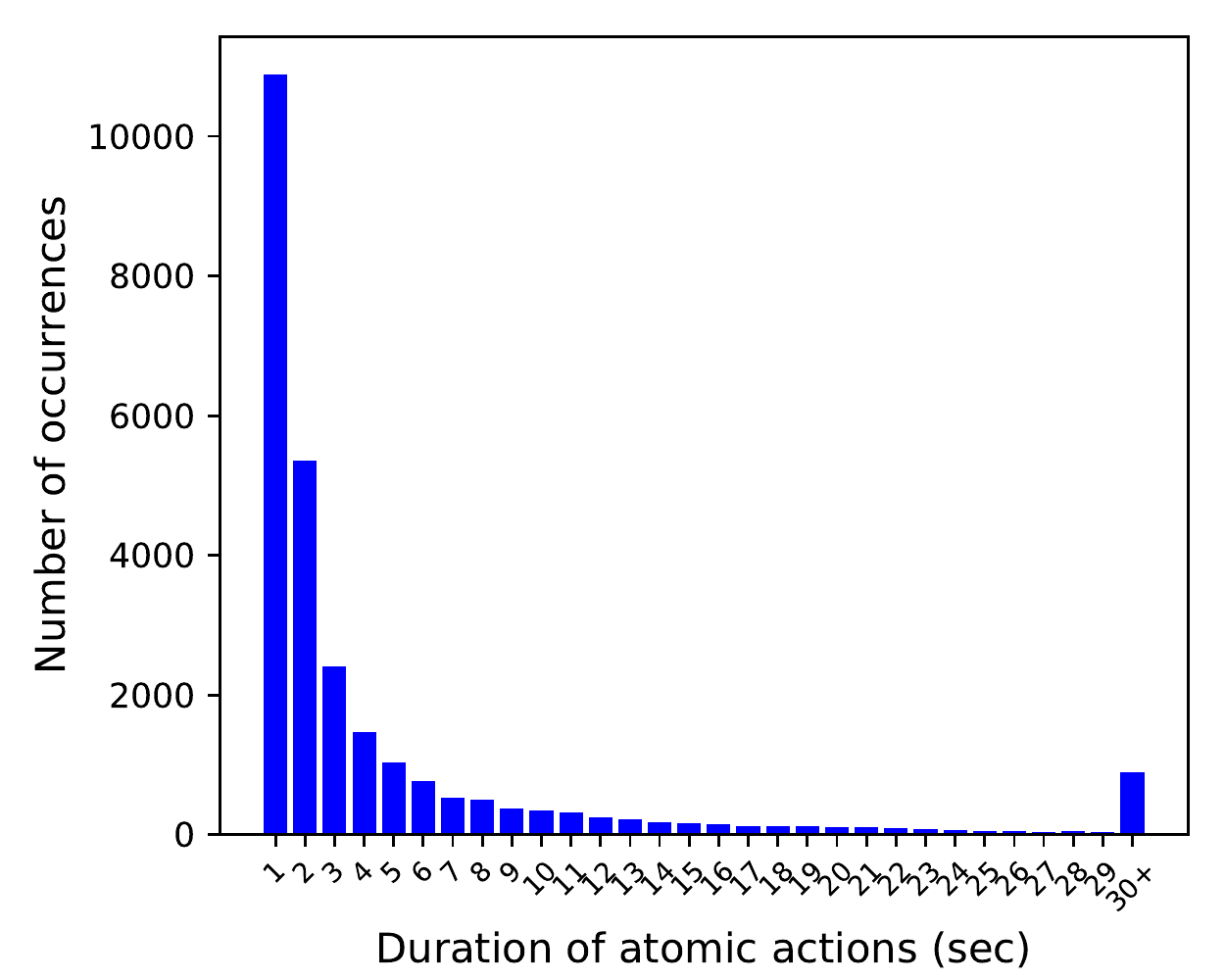}
    \caption{Distribution of duration of atomic actions. \label{fig:dur_dist}}
    \end{center} 
\end{figure}

In this section, we include further details about the HOMAGE dataset. For the spatio-temporal scene graph, Figure \ref{fig:obj_dist} shows the most frequent object classes and Figure \ref{fig:rel_dist} shows the most frequent object relationships. Figure \ref{fig:matrix_obj_rel} shows the joint distribution of object classes and relationships. Figure \ref{fig:dur_dist} shows the distribution of the durations of atomic actions.

To encompass activities in the living space, the types of activities in this dataset were determined by referring to the American Time Use Survey (ATUS), which is a survey of time at home allowing researchers to look at how much time people spend doing different activities. As there are several existing references defining atomic action for daily activities, we borrow definitions from datasets such as Charades \cite{charades}, EPIC-KITCHEN \cite{epic} and Action Genome \cite{action_genome}.

\renewcommand\thefigure{B.\arabic{figure}}    
\setcounter{figure}{0}   
\renewcommand\thetable{B.\arabic{table}}    
\setcounter{table}{0}   
\section*{Appendix B - Additional Experiments}
\label{sec:experiments}

\subsubsection*{Self-Supervised Pre-Training}

\textbf{Approach} Our base backbone remains similar to the one we discuss in the main paper and the overall approach is inspired by \cite{dpc}. To summarize, an aggregation function, $g(\cdot)$ takes a sequence $\{z_1, z_2, \ldots, z_j\}$ as input and generates a context representation $c_j = g(z_1, z_2, \ldots, z_j)$. In our setup, $z_j \in \mathbb{R}^{H' \times W' \times D}$ and $c_j \in \mathbb{R}^{D}$. $D$ represents the embedding size and $H'$, $W'$ represent down-sampled resolutions as different regions in $z_j$ represent features for different spatial locations. We define $z'_j = Pool(z_j)$ where $z'_j \in \mathbb{R}^{D}$ and $c = F(V)$ where $F(\cdot) = g(f(\cdot))$. In our experiments, $H'=4, W'=4, D=256$.

\begin{figure}[t]
      \centering
        \includegraphics[width=\linewidth]{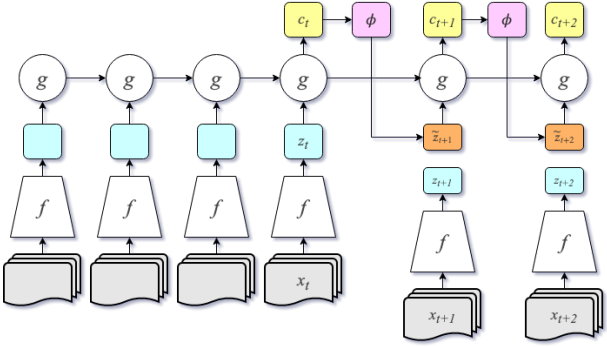}
        \caption{A diagram of the learning framework utilized. We look at features in a sequential manner while simultaneously trying to predict representations for future states.}
        \label{learningFramework}
\end{figure}

To learn effective representations, we create a prediction task involving predicting $z$ of future blocks similar to \cite{dpc}. In the ideal scenario, the task should force our model to capture all the necessary contextual semantics in $c_t$ and all frame-level semantics in $z_t$. We define $\phi(\cdot)$ which takes as input $c_t$ and predicts the latent state of the future frames. The formulation is given in Eq. \eqref{pred-eq}. Fig. ~\ref{learningFramework} provides a compact visual representation of the learning framework.

\begin{equation}
    \begin{split}
        \widetilde{z}_{t+1} = \phi(c_t),\\
        \widetilde{z}_{t+1} = \phi(g(z_1, z_2, \ldots, z_t)),\\
        \widetilde{z}_{t+2} = \phi(g(z_1, z_2, \ldots, z_t, \widetilde{z}_{t+1})),
    \end{split}
    \label{pred-eq}
\end{equation}
where $\phi(\cdot)$ takes $c_t$ as input and predicts the latent state of the future frames. We then utilize the predicted $\widetilde{z}_{t + 1}$ to compute $\widetilde{c}_{t+1}$. We can repeat this for as many steps as we want, in our experiments we restrict ourselves to predict till 3 steps in to the future.
 
Note that we use the predicted $\widetilde{z}_{t+1}$ while predicting $\widetilde{z}_{t+2}$ to force the model to capture long-range semantics. We can repeat this for a varying number of steps, although the difficulty increases tremendously as the number of steps increases as seen in \cite{dpc}. In our experiments, we predict the next three blocks using the first five blocks.

\paragraph{Results} To study the value of multiple viewpoints of the video data, we perform pre-training with the above learning framework weights to get a self-supervised initialization for our experiment. We first train our model in the self-supervised setting for 500 epochs. We use the pre-trained weights to initialize the ego-view and third-person view encoders and train with supervision loss to the same number of epochs as the randomly initialized baseline. Note that in the supervision phase, each modality is trained separately and no cross-modality loss is used. Table \ref{table:self_sup} shows that cooperative learning with different modalities results in distinctively improved performance compared to random initialization as we are able to utilize structural information naturally present in the examples. We also observe that the model with self-supervised pre-training converges faster than the baseline. This demonstrates the additional possibility of utilizing Home Action Genome to evaluate multi-modal self-supervision approaches.

\begin{table}[t]
  \centering
    \begin{tabular}{c|c|c} 
     \toprule
     Method & Ego-View & 3$^{rd}$ Person\\
     \midrule
     SV & 31.8 & 21.8 \\
     SS + SV & \textbf{33.1} & \textbf{24.8} \\
     \bottomrule
    \end{tabular}
    \captionof{table}{Effect of self-supervised pre-training on atomic action classification. We see considerable performance improvements when initializing our model with pre-training using multi-modal self supervision.}
    \label{table:self_sup}
\end{table}

\subsection*{Baseline with Oracle Scene Graphs}

We provide a baseline for human action classification using oracle scene graphs. This experiment gives a rough reference of the upper bound of action inference using spatio-temporal information.

We represent the ground-truth scene graph input as a matrix $M$ of size $n_{obj} \times n_{rel}$, with $n_{obj}$ and $n_{rel}$ be the number of object and relationship categories, respectively, initialized to be filled with 0. We encode a relationship with object category $s$, and relationship category $r$ by setting $M[s, r]$ to be 1. The input representation is then flattened and fed into an MLP-based encoder.

Table \ref{table:sg} shows the performance of activity classification using ground-truth scene graphs, with the encoding scheme described above. We observe that the modality of the ground-truth scene graph is very informative compared with the other modalities, highlighting the potential for scene graph prediction on human action understanding.

\begin{table}[hbt!]
  \centering
    \begin{tabular}{c|c} 
     \toprule
      Acc1 & Acc3\\
     \midrule
      76.0 & 91.7 \\
     \bottomrule
    \end{tabular}
    \captionof{table}{Classification of activities using ground-truth scene graphs. Results are averaged over the two test splits.}
    \label{table:sg}
\end{table}

\subsubsection*{Multi-Task Loss}

As discussed in Section ~\ref{compose}, we utilize two variants for our multi-task losses. The first is an equally weighed variant where both $\mathcal{L}_a$ and $\mathcal{L}_v$ have the same weights, while the other is similar to the one proposed in \cite{kendall2018multi} utilizing task-dependent uncertainty to automatically weigh losses. The loss is defined as:

\begin{equation}
    \mathcal{L}_{c} = \mathcal{L}_{v} \mathbin{/} \sigma_{v}^2 + \mathcal{L}_{a} \mathbin{/} \sigma_{a}^2 + \log ( \sigma_{v} . \sigma_{a} )
\end{equation}

Where $\sigma_i$ refers to the task dependent uncertainty (aleatoric homoscedastic). Although the latter has shown improved results in numerous settings, we noticed that it led to slower convergence and the performance improvements were not consistent across modalities. For this reason, all results reported utilize the simple equally weighted multi-task loss.

\subsubsection*{Learning Attention}

As mentioned in the main text, we also explore the usage of an attention module that allows auto-learning of associations between different modalities similar to \cite{multisensory2018} which do it for audio and visual modalities. We setup attention in a slightly different manner by predicting weights over the grid. Recall that our features are arranged in a grid of shape $H' \times W'$. We predict $H' \times W'$ values $\alpha_{i, j}$ representing the weight of each feature corresponding to spatial location $(i, j)$. Given an original context $c$ of shape $D \times H' \times W'$, we extract $c_{agg}$ from it as given in Eq.~\eqref{attention}. Note that we generate attention weights for each pair of modalities to capture the associations between them.

In our experiments, we did not notice any differences between choosing various values of temperature as it seems the network modulated the learned $\alpha$'s accordingly. $p$'s are utilized to infer regions of interest, as cells with higher $p$ correspond to relevant portions of the modalities. Another thing worth noting is that this attention module is only used in conjunction with image modalities, as we found attention over an audio spectrogram was not directly interpretable in the traditional sense.

\begin{figure}[t]
\begin{center}
    \centering
    \includegraphics[width=\linewidth]{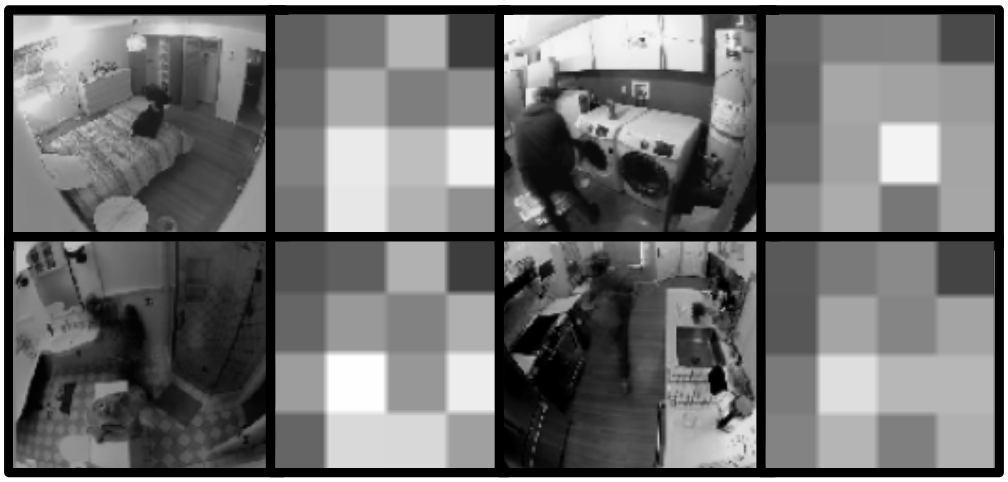}
    \captionof{figure}{\small{Visual results for multi-modal attention between ego-centric and third person view. We show four instances where the left image refers to the third person view, while the right shows the predicted attention weights (White represents higher importance for attention). As we can see, CCAU is loosely able to predict areas of interest using our proposed self-supervised losses.}}
    \label{intuition}
    \centering
\end{center}
\end{figure}

\subsubsection*{Knowledge Distillation}

We discuss Knowledge Distillation briefly in the main text as one of the important baselines in Section 5.3.1. The framework we used is similar to the famously used one proposed in \cite{hinton2015distilling}. Without going into details, the overall loss is given in Eq.~\eqref{kd}.

\begin{equation}
    \label{kd}
    \mathcal{L}_{kd} = \alpha \cdot \mathcal{H}(y, \sigma(zs)) + \beta \cdot \mathcal{H}(\sigma(zt, \tau), \sigma(zs, \tau))
\end{equation}

Eq.~\eqref{kd} is an instance of matching logit distributions leading to the distillation of knowledge from the teacher to the student. Where $H$ represents the cross-entropy loss, $\tau$ represents the temperature. $zs$ and $zt$ are outputs for the student and teacher, respectively.

For multiple modalities, the loss is just repeated multiple times for each modality. For our experiments we use $\alpha=1$ and $\beta=0.1$. We choose $\tau=2.5$ as the models are similar in capacity. We also experiment with two variants i.e. Static and Cooperative Knowledge Distillation. The difference being Static KD involves static teachers while the cooperative variants allow all modalities to serve as both students and teachers.

\end{document}